\documentclass[11pt]{article}

\usepackage[pagebackref,colorlinks,breaklinks=true]{hyperref}
\usepackage{amsmath} 
\usepackage{amsthm} 
\usepackage{amssymb}	
\usepackage{graphicx} 
\usepackage{multicol} 
\usepackage{multirow}
\usepackage{color}
\usepackage[dvips,letterpaper,margin=1in,bottom=1in]{geometry}
\usepackage[capitalize,noabbrev]{cleveref}

\usepackage{bbm}

\usepackage[utf8]{inputenc}
\usepackage[english]{babel}

\usepackage{mathtools}

\makeatletter
\newenvironment{breakablealgorithm}
  {
   \begin{center}
     \refstepcounter{algorithm}
     \hrule height.8pt depth0pt \kern2pt
     \renewcommand{\caption}[2][\relax]{
       {\raggedright\textbf{\ALG@name~\thealgorithm} ##2\par}%
       \ifx\relax##1\relax 
         \addcontentsline{loa}{algorithm}{\protect\numberline{\thealgorithm}##2}%
       \else 
         \addcontentsline{loa}{algorithm}{\protect\numberline{\thealgorithm}##1}%
       \fi
       \kern2pt\hrule\kern2pt
     }
  }{
     \kern2pt\hrule\relax
   \end{center}
  }
\makeatother

\newtheorem{assumption}{Assumption}
\newtheorem{fact}{Fact}
\newtheorem{datainput}{Data Input}
\newtheorem{theorem}{Theorem}[section]

\newtheorem{lemma}[theorem]{Lemma}

\newtheorem{remark}{Remark}[section]

\DeclarePairedDelimiter\ket{\lvert}{\rangle}
\DeclarePairedDelimiter\bra{\langle}{\rvert}

\def\be{\begin{eqnarray}}
\def\ee{\end{eqnarray}}

\usepackage{enumerate}

\usepackage{algorithm}
\usepackage{algorithmic}

\usepackage{tabularx}
\usepackage{booktabs}
\usepackage{threeparttable}

\newcommand{\footremember}[2]{%
    \footnote{#2}
    \newcounter{#1}
    \setcounter{#1}{\value{footnote}}%
}

\usepackage{tikz}
\usetikzlibrary{quantikz2}

\title{Quantum Algorithm for Sparse Online Learning \\ with Truncated Gradient Descent}

\author{Debbie Lim \footremember{1}{Debbie Lim is with the Center for Quantum Computing Science, University of Latvia (email: \href{mailto:limhueychih@gmail.com}{\nolinkurl{limhueychih@gmail.com}}).}
\and Yixian Qiu \footremember{2}{Yixian Qiu is with the Centre for Quantum Technologies, National University of Singapore (email: \href{mailto:yixian_qiu@u.nus.edu}{\nolinkurl{yixian_qiu@u.nus.edu}}).}
\and Patrick Rebentrost \footremember{3}{Patrick Rebentrost is with the Centre for Quantum Technologies, National University of Singapore (email: \href{mailto:cqtfpr@nus.edu.sg}{\nolinkurl{cqtfpr@nus.edu.sg}}).}
\and Qisheng Wang \footremember{4}{Qisheng Wang is with the School of Informatics, University of Edinburgh (e-mail: \href{mailto:QishengWang1994@gmail.com}{\nolinkurl{QishengWang1994@gmail.com}}).}
}
\date{}

\begin{document}

\maketitle

\begin{abstract}
Logistic regression, the Support Vector Machine (SVM), and least squares are well-studied methods in the statistical and computer science community, with various practical applications. High-dimensional data arriving on a real-time basis makes the design of online learning algorithms that produce sparse solutions essential. The seminal work of \hyperlink{cite.langford2009sparse}{Langford, Li, and Zhang (2009)}
developed a method to obtain sparsity via truncated gradient descent, showing a near-optimal online regret bound. Based on this method, we develop a quantum sparse online learning algorithm for logistic regression, the SVM, and least squares. Given efficient quantum access to the inputs, we show that a quadratic speedup in the time complexity with respect to the dimension of the problem is achievable, while maintaining a regret of $O(1/\sqrt{T})$, where $T$ is the number of iterations. 
\end{abstract}

\newpage

\tableofcontents
\newpage

\section{Introduction}
The field of statistical modeling and machine learning is about developing robust methodologies to uncover patterns, make predictions, and derive insights from data. Three prominent techniques are logistic regression, support vector machines (SVMs), and least squares. Regression and data fitting make use of least squares~\cite{hansen2013least, eberly2000least, cantrell2008review, watson1967linear, geladi1986partial, audibert2011robust, maillard2009compressed, boyd2018introduction} to study the relationship between predictor variables and a response variable in a set of data. In particular, given $N$ data points and their labels  $\{x^{(i)}, y^{(i)}\}_{i=1}^N$ such that $x^{(i)}\in\mathbb R^d$ and $y^{(i)}\in\mathbb R$ for all $i\in [N]$\footnote{$[N]$ denotes the set $\{1, \cdots, N\}$.}, and a model function $f:\mathbb R^d\rightarrow \mathbb R$, the goal is to find the optimal $w$ that minimizes the squared loss $ \sum_{i=1}^N (f\left(x^{(i)}, w\right) -y^{(i)})^2$. 

Unlike linear regression which models continuous outcomes, logistic regression is adept at predicting the probability of a discrete outcome, for example success or failure, making it an essential tool for understanding and predicting categorical data~\cite{lavalley2008logistic, nick2007logistic, menard2002applied, das2021logistic, sperandei2014understanding, stoltzfus2011logistic}. In short, logistic regression can be described as follows: given a set of $N$ data points and their labels $\{x^{(i)}, y^{(i)}\}_{i=1}^N$ such that $x^{(i)}\in\mathbb R^d$ and $y^{(i)}\in\{-1, 1\}$ for all $i\in[N]$, logistic regression aims to find a $w\in\mathbb R^d$ that minimizes the loss $\sum_{i=1}^N \ln (1 + e^{-(w\cdot x^{(i)})\cdot y^{(i)} })$.   

On the other hand, SVM is a widely used tool in machine learning and finds applications in the domain of chemistry, biology, finance~\cite{ivanciuc2007applications, huang2018applications, yang2004biological, tay2001application}, due to its simplicity of use and robust performance. A support vector machine is an algorithm that classifies vectors in a feature space into one of two sets, given training data from the sets~\cite{cortes1995support}. SVM works by constructing the optimal hyperplane that partitions the two sets, either in the original feature space or a higher-dimensional, or even infinite dimensional, kernel space. In the soft-margin setting of the SVM, given a set of $N$ data points and their labels $\{x^{(i)}, y^{(i)}\}_{i=1}^N$ such that $x^{(i)}\in\mathbb R^d$ and $y^{(i)}\in\{-1, 1\}$ for all $i\in[N]$, the goal is to find a $w\in\mathbb R^d$ that minimizes the hinge loss $\sum_{i=1}^N \max\{0, (1 - y^{(i)}w \cdot x^{(i)})\}$.

\paragraph{The online learning framework.} Online learning algorithms have gained much attention in recent decades, in both the academic and industrial sectors~\cite{hoffman2010online, eon1998online, kivinen2004online, dekel2012optimal, helmbold1998line, anava2013online, hoi2021online}. In this framework, the learner (also known as the learning algorithm) who is given access to partial, sequential training data, is required to output a solution based on partial knowledge of the training data. The solution is then updated in the next iteration after receiving more training data as input. This process is repeated for $T$ number of iterations. More specifically, for every time $t=1, \cdots, T$, the following sequence of events take place: 
\begin{enumerate}[1)]
    \item The learner receives an unlabelled example $x^{(t)}$;
    \item The learner makes a prediction $\hat y^{(t)}$ based on an existing weight vector $w^{(t)}\in\mathbb R^d$; 
    \item The learner receives the true label $y^{(t)}$ and suffers a loss $L(w^{(t)}, x^{(t)}, y^{(t)})$ that is convex in $w^{(t)}$;
    \item The learner updates the weight vector according to some update rule $w^{(t+1)}\gets f(w^{(t)})$. 
\end{enumerate}

Due to the fact that the input data of online algorithms can be adversarial in nature, such algorithms are particularly useful in proving guarantees for worst-case inputs. Besides having an efficient running time, the design of online algorithms focuses on regret minimization. The \emph{regret} of an online algorithm is defined as the difference between the total loss incurred using a certain sequence of strategies and the total loss incurred using the best fixed strategy in hindsight \cite{hazan2007logarithmic}. Specifically,\footnote{Strictly speaking, this is the per-step regret as we normalize by $T$. While the conventional regret is the unnormalized version, we nevertheless call this the regret in this paper.} 
\[
\mathit{Regret} \coloneqq \frac{1}{T}\sum_{t=1}^T L\left(w^{(t)}, x^{(t)}, y^{(t)}\right) - \min_{u\in\mathbb R^d}\frac{1}{T}\sum_{t=1}^T L\left(u, x^{(t)}, y^{(t)}\right).
\]

In the era of big data, we are often faced with large and high-dimensional problem data. As a result, the solution to the learning problem will inherit a large dimension. Techniques such as best subsets, forward selection, and backward elimination~\cite{mao2002fast, mao2004orthogonal, whitley2000unsupervised, ververidis2005sequential, borboudakis2019forward, reif2014efficient, tan2008genetic, zongker1996algorithms, kumar2014feature, wei2006feature, dai2018some, peng2022portfolio, zhang2015sparse} enhance computational efficiency and ease the interpretability of the solution. Seeing the importance of sparse solutions and the strength of online algorithms, the need for sparse online learning is apparent~\cite{liang2021screening, wang2015framework, lin2016sparse}. 

The work of \cite{langford2009sparse} introduced a truncated gradient descent algorithm for sparse online learning. In their algorithm, the solution is updated via gradient descent at every iteration and a truncation is performed on the solution after every $K$ iterations, where $K$ needs to be carefully chosen (see Algorithm~\ref{algo} in  Appendix~\ref{app:algo}). In their work, the following assumptions are made. 
\begin{assumption}[\cite{langford2009sparse}]\label{ass: regret_ass}For every $t\in[T]$, 
    \begin{enumerate}[(i)]
        \item  The loss function $L(w^{(t)}, x^{(t)}, y^{(t)})$ is convex in $w^{(t)}$ for all $x^{(t)}, y^{(t)}$. 
        \item There exist constants $A, B\in\mathbb R_{>0}$ such that $\Vert \nabla_{w^{(t)}} L(w^{(t)}, x^{(t)}, y^{(t)})\Vert_2^2\leq A \cdot L(w^{(t)}, x^{(t)}, y^{(t)}) + B$  for all $x^{(t)}, y^{(t)}$, where $\Vert\cdot\Vert_2$ denotes the Euclidean norm. 
        \item \label{ass:sup_x} $\sup_{x^{(t)}} \lVert x^{(t)} \rVert_2 \leq C$ for some constant $C\in\mathbb R_+$. 
    \end{enumerate}
\end{assumption}
Under these assumptions, the authors of \cite{langford2009sparse} showed that their online algorithm achieves an $O(1/\sqrt T)$ regret (refer to Fact~\ref{fact:classical_regret} in Appendix~\ref{app:proof_regret}). 
As also noted in \cite{langford2009sparse}, the general loss function for linear prediction problems is of the form $L(w^{(t)}, x^{(t)}, y^{(t)}) = h(w^{(t)T}x^{(t)}, y^{(t)})$. They pointed out some common loss functions $h(\cdot, \cdot)$ with corresponding choices of parameters $A$ and $B$ (which are not necessarily unique), under the assumption that $\sup_{x^{(t)}} \lVert x^{(t)} \rVert_2 \leq C$. Among them are 
\begin{itemize}
    \item Logistic regression: $h(w^{(t)T}x^{(t)}, y^{(t)}) = \ln (1 + \exp(-w^{(t)T}x^{(t)}\cdot y{(t)}))$; $A = 0$, $B = C^2$, $y^{(t)}\in\{\pm 1\}$ for all $t\in[T]$. 
    \item SVM (hinge loss):  $h(w^{(t)T}x^{(t)}, y^{(t)}) = \max\{0, 1 - w^{(t)T}x^{(t)}\cdot y{(t)}\}$; $A = 0$, $B = C^2$, $y^{(t)}\in\{\pm 1\}$ for all $t\in[T]$. 
    \item Least squares (square loss): $h(w^{(t)T}x^{(t)}, y^{(t)}) = (w^{(t)T}x^{(t)} - y^{(t)})^2$; $A = 4C^2$, $B = 0$, $y^{(t)}\in\mathbb R$  for all $t\in[T]$. 
\end{itemize}

\paragraph{Main contribution.} In this work, we present a quantum online algorithm that outputs a sparse solution, which has applications to logistic regression (Section \ref{Quantum algorithm:LR}), the SVM (Section \ref{Quantum algorithm:SVM}) and least squares (Section \ref{Quantum algorithm:LS}). Our work is based on \cite{langford2009sparse}, who introduced a truncated gradient descent algorithm for sparse online learning. The guarantees of our algorithm hold under the same assumptions as~\cite{langford2009sparse} (to be discussed later). While maintaining the $O(1/\sqrt T)$ regret bound of \cite{langford2009sparse}, our quantum algorithm has time complexity of $\tilde O(T^{5/2}\sqrt d)$,\footnote{$\tilde O(\cdot)$ suppresses polylogarithmic factors.} achieving a quadratic speedup in the dimension over the classical $O(Td)$, where $d$ is the dimension of a data point. This speedup is noticeable when $d\geq \tilde\Omega(T^5)$, making the algorithm useful for high-dimensional learning tasks. We summarize our results in the Table~\ref{tab:my_label}. 

\begin{table}[!htp]\label{tab:my_label}
    \centering
    \caption{Summary of results}
    \begin{tabular}{l l l l l}
    \toprule
    \multirow{2}{*}{\textbf{Problem}} & \multicolumn{2}{c}{\textbf{Time Complexity}} & \multicolumn{2}{c}{\textbf{Regret}} \\
    \cmidrule(lr){2-3} \cmidrule(lr){4-5}
    & \cite{langford2009sparse} & \textbf{Our Work} & \cite{langford2009sparse} & \textbf{Our Work} \\
    \midrule
    Logistic regression  & $O(Td)$ & $\tilde O(T^{5/2}\sqrt{d})$ & $O(1/\sqrt T)$ & $O(1/\sqrt T)$ \\
    SVM                & $O(Td)$ & $\tilde O(T^{5/2}\sqrt{d})$ & $O(1/\sqrt T)$ & $O(1/\sqrt T)$ \\
    Least squares       & $O(Td)$ & $\tilde O(T^{5/2}\sqrt{d})$ & $O(1/\sqrt T)$ & $O(1/\sqrt T)$ \\
    \bottomrule
    \end{tabular}
    \label{tab:summary_results}
\end{table}

Our algorithm does not need to read in all the entries of the input data at once.
For a data point $x^{(t)} \in \mathbb{R}^d$, the $j$-th entry of $x^{(t)}$, $x^{(t)}_j$, can be accessed in $\tilde O(1)$ time on a classical computer. 
We assume that $x^{(t)}_j$ can be accessed in $\tilde O(1)$ time coherently on a quantum computer (formally defined in Data Input~\ref{data_input}), which is the \textit{standard quantum input model} employed in the previous literature, e.g.,~\cite{grover1996fast, li2019sublinear, brandao2019quantum}.

Our quantum algorithm returns the weight vectors $w^{(1)}, w^{(2)}, \dots, w^{(T)}$ indirectly. 
Specifically, for each $w^{(t)}$ with $1 \leq t \leq T$, our algorithm enables us to \textit{coherently} access each of its entries in $O(t)$ time. 
Notably, the cost of accessing one entry of $w^{(t)}$ is upper bounded by $O(T)$, and $T$ is usually set to $O(1/\epsilon^2)$ (which is \textit{independent} of the dimension $d$) if we want a regret of $\epsilon$ in concrete applications such as logistic regression, SVM, and least squares in Table~\ref{tab:my_label}.
Especially for constant regret, e.g., $\epsilon = 0.1$, the cost is a constant time.
In addition, this output model is even useful when we do not need to know the exact value of each entry of $w^{(t)}$ but certain expectations with respect to $w^{(t)}$. 
As suggested in \cite{harrow2009quantum}, the (normalized) quantum state $\ket{w^{(t)}}$ can be useful in estimating the expectations.
To this end, our quantum algorithm allows us to further prepare $\ket{w^{(t)}}$ with an extra time complexity of $\tilde O(t\sqrt{d})$ through the standard quantum state preparation \cite{grover2000synthesis}; this is negligible compared to the overall time complexity $\tilde O(T^{5/2}\sqrt{d})$.

\paragraph{Techniques.} Our quantum algorithm is based on the framework of \cite{langford2009sparse}. In this framework, the algorithm maintains a sparse weight vector by performing the truncation regularly. Our main observation is that in the framework, a large number of updates (linear in the data dimension $d$) are required (on a classical computer) while the prediction in each iteration is just a single real number.
This motivates us to find a reasonable trade-off between the update and the prediction, and we then realized how to achieve this on a quantum computer. 

Our quantum speedup comes from the techniques that rely on quantum amplitude estimation and amplification~\cite{brassard2002quantum, harrow2020adaptive, rall2023amplitude, cornelissen2023sublinear}. In particular, we use subroutines such as quantum inner product estimation, quantum norm estimation and quantum state preparation. This allows us to obtain a quadratic speedup in the dimension $d$ for the prediction. 
For the update, we do not actually perform the updates but implement them in an oracle-oriented manner so that any entry of the intermediate vectors can be computed in $\tilde O(T)$ time, which is sufficient for us to make the prediction efficiently on a quantum computer.
Specifically, we leverage the circuits for efficient arithmetic operations to avoid storing the weight vector in every iteration, thereby saving the space and time of the algorithm.
Under this quantized framework, we develop quantum algorithms for logistic regression, the SVM and least squares by specifying the quantum circuits with appropriate parameters for the corresponding 

\paragraph{Applications.}

Let $u^*$ be the best fixed strategy in hindsight. For logistic regression and the SVM, we observe that by taking $T = \Theta(C^4\Vert u^*\Vert_2^4/\epsilon^2)$, the regret of our quantum algorithm becomes $\Theta(\epsilon)$.
This setting implies quantum algorithms for (offline) logistic regression and SVM with time complexity $\tilde O(C^{10} \Vert u^*\Vert_2^{10}\sqrt d /\epsilon^5)$, which achieves an exponential improvement in the dependence on $C\Vert u^*\Vert_2$ for logistic regression and a polynomial improvement in the $\epsilon$-dependence for the SVM, compared to the prior best offline results due to \cite{shao2019fast,li2019sublinear} respectively. Moreover, taking $T = \Theta((C^6 + C^4 \Vert u^*\Vert_2^4)/\epsilon^2)$ results in a $\Theta(\epsilon)$ regret for least squares and a time complexity of $\tilde O(C^{15} \Vert u^*\Vert_2^6\sqrt d /\epsilon^5)$, if the prediction error is constant-bounded. This setting implies a quantum algorithm for (offline) least squares. For reference, a quantum algorithm for offline least squares with different conditions was presented in \cite{liu2017fast}.\footnote{In these offline settings, the output is the ``average'' vector $\bar w = \frac{1}{T}\sum_{t=1}^T w^{(t)}$. For a fair comparison with the offline results, after the execution of our algorithm, it provides quantum access to each entry of $\bar w$ at an extra cost of $\tilde O(T^2)$, where $T$ is usually chosen to be independent of the dimension $d$ in the presented applications.}

\section{Related Work}
Motivated by the importance of obtaining sparse solutions for the aforementioned reasons in the previous section,
various methods have been used to achieve sparsity in learning algorithms. Some examples include randomized rounding~\cite{golovin2013large}, forward sequential selection~\cite{cotter2005sparse}, backward sequential elimination~\cite{cotter2001backward} and adaptive forward-backward greedy algorithm~\cite{zhang2008adaptive}. 

In the online setting, a method called Forward-Backward Splitting (FOBOS) with an $\ell_1$-norm regularizer has been proposed by Duchi \textit{et al.}~\cite{duchi2009efficient}. This approach is analogous to the projected gradient descent, where the projection step is replaced with a minimization 
problem. Inspired by dual-averaging techniques, the Regularized Dual Averaging (RDA) algorithm by Xiao \textit{et al.}~\cite{xiao2009dual} is designed for problems whose objective function consists of a convex loss function and a convex  regularization term. Under certain conditions, it achieves $O(1/\sqrt{T})$ regret and proves effective for sparse online learning with $\ell_1$-regularization. The authors in \cite{wang2013online} gave two online feature selection algorithms, which are modifications of the Perceptron algorithm~\cite{rosenblatt1958perceptron}. The performance of both algorithms is evaluated numerically and via the mistake bound. Other references on sparse online learning include~\cite{zhao2020unified, mairal2010online, ma2017stabilized, song2019online, hao2021online, liu2019adaptive}. 

In the offline setting, quantum algorithms for least squares and SVM have been studied. The recent work by \cite{song2023revisiting} gave a quantum algorithm that outputs a solution such that the $\ell_2$-norm of the residual vector approximates that of the optimal solution up to a relative error of $\epsilon > 0$ with high probability. Their algorithm runs in time $\tilde O(\sqrt n d^{1.5}/\epsilon + d^\omega/\epsilon)$, where $n$ is the sample complexity and $\omega\approx 2.37$ denotes the exponent of matrix multiplication. This improves upon the best classical algorithm which runs in time $O(nd) + \operatorname{poly}(d/\epsilon)$~\cite{clarkson2017low}. Moreover, Liu and Zhang \cite{liu2017fast} proposed a quantum algorithm that solves the same problem in time $O\left(\log(n+d) s^2\kappa^3/\epsilon^2\right)$, where $s$ denotes the sparsity of the data matrix and $\kappa$ is the condition number. Other quantum algorithms for linear regression \cite{wang2017quantum, kerenidis2016quantum, chakraborty2018power} have been discussed. However, the time complexity of these algorithms depends on some quantum linear-algebra \cite{Zhao2021_QLA} related parameters, such as the condition number of the data matrix \cite{wang2017quantum, kerenidis2016quantum, chakraborty2018power}. For the SVM, the authors in \cite{rebentrost2014quantum} showed a quantum algorithm based on HHL and quantum linear algebra techniques. For the case of general data sets, Li, Chakrabarti and Wu \cite{li2019sublinear} gave a quantum algorithm that runs in time $\tilde O\left(\frac{\sqrt n}{\epsilon^4} + \frac{\sqrt d}{\epsilon^8}\right)$, improving over the classical running time of $O\left(\frac{n+d}{\epsilon^2}\right)$ by \cite{clarkson2012sublinear}. The complexity of the quantum SVM has been studied by \cite{gentinetta2024complexity}. Several sublinear time quantum algorithms were developed under the framework of online learning, e.g., for semidefinite programming \cite{brandao2019quantum,van2020quantum,brandao2017quantum,VanApeldoorn2019}, zero-sum games \cite{li2019sublinear,van2019quantum,bouland2023quantum,gao2024logarithmic}, general matrix games \cite{li2021sublinear} and learning of quantum states~\cite{aaronson2018online,Yang20aaai,Chen24quantum}.

With regards to how quantum computing can improve the efficiency of algorithms on feature selection, Saeedi and Arodz~\cite{saeedi2019quantum} proposed the Quantum Sparse Support Vector Machine (QsSVM), an approach that minimizes the training-set objective function of the Sparse SVM model~\cite{bennett1999combining, kecman2000support, bi2003dimensionality, zhu20031} by using a quantum algorithm for solving linear programs (LPs)~\cite{van2019quantum} instead of a classical LP solvers. While quantum LP solvers may not speed up arbitrary binary classifiers, they offer sublinear time complexity in the number of samples and features for sparse linear models, unlike classical algorithms. 
Sampling can be considered a form of feature selection. The quantum online portfolio optimization algorithm by \cite{lim2024quantum} employs quantum multi-sampling to invest in sampled assets. Their approach achieves quadratic speedup compared to classical methods~\cite{helmbold1998line}, with a marginal increase in regret. The authors of~\cite{lin2020quantum} studied quantum-enhanced least-square SVM with two quantum algorithms. The first employs a simplified quantum approach using continuous variables for matrix inversion, while the second is a hybrid quantum-classical method providing sparse solutions with quantum-enhanced feature maps, both achieving exponential speedup in sample size. Other related quantum algorithms include~\cite{liu2019quantum, wang2019quantum, rebentrost2014quantum, doriguello2023quantum, liu2017fast}.  

\section{Preliminaries}\label{sec:preliminaries}
\paragraph{Notations.} For a positive integer $d\in\mathbb Z_+$, we use $[d]$ to represent the set $\{1, \cdots, d\}$. Given a vector $u\in\mathbb R^d$, we denote the $j$-th entry of $u$ as $u_j$ for all $j\in[d]$ and denote the $\ell_1$-norm and $\ell_\infty$ norm of $u$ as $\lVert u\rVert_1 := \allowdisplaybreaks\sum_{j=1}^d \vert u_j\vert$ and $\lVert u\rVert_\infty: = \max_{j\in[d]} \vert u_j\vert$. If the vector has a time dependency, we denote it as $u^{(t)}$. For some condition $C$, we use $I(C)$ to denote the indicator function that evaluates to 1 if $C$ is satisfied and 0 otherwise. We use $\bar 0$ to denote the all zeros vector and use $\ket{\bar 0}$ to denote the state $\ket{0}\otimes \cdots\otimes \ket{0}$ where the number of qubits is clear from the context.  We use $\tilde O(\cdot)$ to hide polylog factors, i.e. $\tilde O(f(n, m)) = O(f(n, m)\cdot \operatorname{polylog}(n, m))$. 

\paragraph{Quantum computational model.}
A quantum algorithm is described by a quantum circuit with queries to the input oracle.
We define the query complexity of a quantum algorithm as the number of queries to the input oracle. 
The time complexity of a quantum algorithm is the sum of its query complexity and the number of elementary quantum gates in it. We assume a quantum arithmetic model, which allows us to ignore issues arising from the fixed-point representation of real numbers. In this model, each elementary arithmetic operation takes a constant time. Our quantum algorithm assumes quantum query access to the input oracle for certain vectors. 
For a vector $u\in\mathbb R^d$, the input oracle for $u$ is a unitary operator $O_u$ such that $O_u \ket{j} \ket{\bar 0} = \ket{j} \ket{u_j}$ for every $j \in [d]$, where the second register is assumed to contain sufficiently many qubits to ensure that all subsequent computations are accurate, in analogy to the sufficient bits that a classical algorithm assumes to run correctly. 

\paragraph{Truncated gradient descent.}
One of the most popular approaches for minimizing a convex loss function $L$ is gradient descent. Starting from an initial point $w^{(1)}$, gradient descent performs 
\begin{equation*}
w^{(t+1)} = w^{(t)} + \eta^{(t)}\nabla L(w^{(t)})
\end{equation*}
for $t = 1,  2, 3, \cdots$, where $\eta^{(t)}>0$ is the step size/learning rate at iteration $t$ and $\nabla L(w^{(t)})$ is the gradient of $L$ at $w^{(t)}$. The algorithm is summarized in Algorithm~\ref{algo:GD} (refer to Appendix~\ref{app:algo}). Seeing the need for sparse solutions in the high-dimensional regime, the authors of~\cite{langford2009sparse} introduced a truncated gradient descent update rule that truncates each entry $j\in[d]$ of the weight vector after certain number of iterations according to the following function: for some threshold $\theta>0$ and a \emph{gravity} parameter\footnote{The gravity parameter measures the amount of shrinkage.} $\alpha>0$, 
\be\label{eqn:trunc}
    \mathcal T(w_j^{(t)}, \alpha, \theta) = \begin{cases}
        \max\{w_j^{(t)} - \alpha, 0\}, \text{ if } 0\leq w^{(t)}_j\leq \theta\\
        \min\{w_j^{(t)} + \alpha, 0\}, \text{ if } -\theta \leq w^{(t)}_j\leq 0\\
        w^{(t)}_j, \text{ otherwise.}\\
    \end{cases}
\ee
The truncated gradient descent method is summarized in Algorithm~\ref{algo:TGD} (refer to Appendix~\ref{app:algo}).

While adaptive learning rates could potentially speed up the convergence of gradient descent~\cite{grimmer2023provably, zeiler2012adadelta, malitsky2023adaptive},  we adopt a constant learning rate with fixed $\eta >0$ as in the setting of \cite{langford2009sparse} for simplicity. Moreover, while the choice of gravity parameters is usually kept open in practice, we shall only consider the following choice: for all $t\in[T]$, 
$\alpha^{(t)} = g^{(t)}\eta$ such that $g^{(1)} = \cdots = g^{(T)} \leq  g_{\max}$, where $g_{\max}$ is some constant. 

\section{Quantum Algorithm}\label{sec:aquantum_algorithm}
In this section, we present our quantum algorithm for sparse online learning, which has applications to logistic regression, the SVM and least squares. We clarify the data input and output model, as well as quantum subroutines in the next subsections. 

\subsection{Quantum input and output model}\label{sec:data_input}
We assume quantum query access via an oracle for the entries of unlabelled examples. The online nature of the problem is given by the fact that we obtain these oracles at different times. 

\begin{datainput}[Online example oracles]\label{data_input}
Let $x^{(1)}, \cdots, x^{(T)}\in\mathbb R^d$ be unlabelled examples. Define the unitaries $U_{x^{(t)}}$ operating on $O(\log d)$ qubits such that for all $j\in[d]$ and $t\in[T]$, $U_{x^{(t)}}\ket{j}\ket{\bar 0} \rightarrow \ket{j}\ket{x^{(t)}_j}$. At time $t\in[T]$, assume access to $U_{x^{(1)}}, \cdots, U_{x^{(t)}}$. 
\end{datainput}

As for the output of our quantum algorithm, it returns weight vectors $w^{(1)}, w^{(2)}, \dots, w^{(T)}$ indirectly. In particular, for all $t\in[T]$, we are allowed coherent access to the entries of $w^{(t)}$ in $O(t)$ time. This cost is upper bounded by $O(T)$. In addition, this output model allows us to further prepare $\ket{w^{(t)}}$, through the standard quantum state preparation \cite{grover2000synthesis}, as suggested in \cite{harrow2009quantum}, such a  (normalized) quantum state $\ket{w^{(t)}}$ can be useful in estimating expectations.

\subsection{Quantum subroutines}
By approximating real numbers to sufficient accuracy, the truncation function Eq.~(\ref{eqn:trunc}) can be efficiently computed by assuming access to \emph{minmax} and \emph{between} oracles. Similar oracles have been studied in ~\cite{vedral1996quantum, ambainis2007quantum, addanki2021design, luongo2024measurement}.

\begin{lemma}[Minmax and Between oracles~\cite{luongo2024measurement}]\label{defn:comparison}
    Let $a, b, x\in\{0, 1\}^n$ and $c, z\in\{0, 1\}$. 
    \begin{enumerate}[(i)]
        \item We say that we have access to a comparison oracle $\mathcal O_{comp}$ if we have access to a unitary $U_{comp}$ that  performs the operation 
        $U_{comp}:\ket{x}\ket{a}\ket{z}\mapsto \ket{x}\ket{a}\ket{z\oplus \mathbbm 1_{x<a}}$
        using $s_{comp}$ Toffoli\footnote{A Toffoli gate as a T-count of 7 and a T-depth of  3.} gates. 
        \item We say that we have access to a controlled-comparison oracle $\mathcal O'_{comp}$ if we have access to a unitary $U'_{comp}$ that  performs the operation $U'_{comp}:\ket{c}\ket{x}\ket{a}\ket{z}\mapsto \ket{c}\ket{x}\ket{a}\ket{z\oplus c\cdot \mathbbm 1_{x<a}}$
        using $s'_{comp}$ Toffoli gates.
        \item Assuming access to a comparison oracle and a controlled-comparison oracle, there exists a circuit that performs the operation $
        U_{Btw}:  \ket{a}\ket{b}\ket{x}\ket{z}\mapsto  \ket{a}\ket{b}\ket{x} \ket{z\oplus \mathbbm 1_{x\in [a, b]}}$
    using $1.5s_{comp} + s'_{comp}$ Toffoli gates. We call this the Between oracle. 
    \end{enumerate}
    The values of $s_{comp}$ and $s'_{comp}$ depend on the type of circuit architecture used for the comparators~\cite{gidney2018halving, cuccaro2004new}. Nevertheless, these are in general $O(n)$.  
    We say that we have access to a minmax oracle if we have access to a unitary $U_{\text{minmax}}$ that performs the following operation
        \begin{align*}
            U_{\text{Minmax}}\ket{c}\ket{x}\ket{0} = \begin{cases} \ket{c}\ket{x}\ket{\max(x, 0)}, \text{ if } c=1\\
            \ket{c}\ket{x}\ket{\min(x, 0)}, \text{ if } c=0.
            \end{cases}
        \end{align*}
        using $O(n)$ number of Toffoli gates. 
\end{lemma}

Using the oracles defined above, we show the following lemma, whose proof can be found in Appendix~\ref{app:truncation_unitary}. 
\begin{lemma}[Truncation unitary]\label{lem:truncation}
    Let $\theta, \alpha\in\mathbb R_{>0}$. Assuming access to a Between oracle and a Minmax oracle, there exists a unitary operator $U_{\mathcal T,  \alpha, \theta}$ that does the following operation up to sufficient accuracy in constant time: $U_{\mathcal T,  \alpha, \theta} \colon \ket{x} \ket{0} \mapsto \ket{x} \ket{f(x)}$,
    where
    \[
        f(x) = \begin{cases}
        \max\{x-\alpha, 0\}, & 0 < x \leq \theta, \\
        \min\{x+\alpha, 0\}, & -\theta \leq x < 0, \\
        x, & \textit{otherwise}.
        \end{cases}
    \]
\end{lemma}
    
Unbiased amplitude estimation~\cite{harrow2020adaptive, rall2023amplitude, cornelissen2023sublinear} allows one to obtain nearly unbiased estimates with low variance and without destroying their initial quantum state. 
\begin{fact}[Unbiased amplitude estimation {\cite[Theorem 2.4]{cornelissen2023sublinear}}]\label{fact:new_AE}\label{fact:noew_AE}
    Let $t \ge 4$ and $\epsilon \in (0,1)$. We are given one copy of a quantum state $\ket \psi$ as input, as well as a unitary transformation $U = 2\ket{\psi}\bra{\psi} - I$, and a unitary transformation $V = I - 2P$ for some projector $P$. There exists a quantum algorithm that outputs $\tilde a$, an estimate of $a = \|P\ket{\psi}\|^2$, such that
    \begin{equation*}
       |\mathbb{E}[\tilde a] - a| \le \epsilon \text{ and } \operatorname{Var}[\tilde a] \le \frac{91 a}{t^2} + \epsilon
    \end{equation*}
    using $\mathcal{O}(t\log \log (t) \log (t/\epsilon))$ applications of $U$ and $V$ each. The algorithm restores the quantum state $\ket \psi$ at the end of the computation with probability at least $1-\epsilon$.
\end{fact}

Quantum state preparation, norm and inner production are widely used subroutines~\cite{van2019quantum, brassard2002quantum, hamoudi2019quantum, li2019sublinear, rebentrost2021quantum} that rely on amplitude estimation~\cite{brassard2002quantum}. We restate these subroutines for the convenience of the reader. The proof is based on Fact~\ref{fact:new_AE}, which is deferred to Appendix~\ref{app:proof_quantum_norm_est_and_quantum_state_prep}. We also note that an additive variant of Lemma~\ref{lem:quantum_norm_est_and_quantum_state_prep}(\ref{lem:quantum_norm_estimation_multiplicative}) can be easily derived with the same time complexity and resources.

\begin{lemma}[Quantum norm estimation and state preparation~\cite{van2019quantum, brassard2002quantum, hamoudi2019quantum, li2019sublinear, rebentrost2021quantum}]\label{lem:quantum_norm_est_and_quantum_state_prep} Let $u\in\mathbb R^d$ and assume quantum access to $u\in\mathbb R^d$. Then,
\begin{enumerate}[(i)]
\item \label{lem:quantum_norm_estimation_multiplicative} Let $\delta\in(0, 1/4)$ and $\epsilon_0 \in (0,1)$. 
There exists a quantum algorithm that outputs an estimate $\tilde \Gamma$ of $\lVert u\rVert_1$ such that $\lvert \tilde \Gamma - \lVert u\rVert_1\vert\leq \epsilon_0\left\Vert u\right\Vert_1$ with probability at least $1-4\delta$. The algorithm runs in time $\tilde O\left(\frac{\sqrt{d}}{\delta }\right)$. 
\item \label{lem:quantum_state_preparation} Let $\zeta\in(0, 1/2]$ and $\tilde \Gamma >0$ be given such that $\vert \lVert u\rVert_1 - \tilde \Gamma\vert \leq \zeta\lVert u\rVert_1$. Let $\delta\in(0, 1)$. An approximation $\vert \tilde p\rangle  = \sum_{j=1}^d \sqrt{\vert p_j\vert}\ket{j}$ of the state $\vert u\rangle = \sum_{j=1}^d \sqrt{\frac{\vert u_j\vert}{\lVert u\rVert_1}}\ket{j}$ can be prepared with probability $1 - \delta$ in $O\left(\sqrt{d}\log\frac{1}{\delta}\right)$ time and $\tilde O\left(\sqrt{d} \log\frac{1}{\zeta}\log\frac{1}{\delta}\right)$ gates. The approximation in $\ell_1$-norm of the probabilities is $\lVert \tilde p - \frac{u}{\lVert u\rVert_1}\rVert_1\leq 2\zeta$
\end{enumerate}
\end{lemma}

The following lemma explains a quantum inner product estimation algorithm which uses amplitude estimation to approximate the inner product between two real vectors. We defer the proof to Appendix~\ref{app_proof_quantum_inner_product_estimation}.

\begin{lemma}[Quantum inner product estimation~\cite{yang2023quantum}]\label{lem:quantum_inner_product_estimation}
Let $\epsilon, \delta\in(0, 1)$ and given access to nonzero vectors $u, v\in\mathbb R^d$. An estimate $\widetilde{IP}$ of the inner product $u\cdot v$ can be obtained such that $\vert \widetilde{IP} - u\cdot v\vert \leq\epsilon$ with success probability at least $1-\delta$. This requires $O\left(\frac{\lVert u\rVert_\infty\lVert v\rVert_1\sqrt{d}}{\epsilon}\log \frac{1}{\delta}\right)$ queries and $\tilde O\left(\frac{\lVert u\rVert_\infty\lVert v\rVert_1\sqrt{d}}{\epsilon}\log \frac{1}{\delta}\right)$ quantum gates. 
\end{lemma}

\subsection{Quantum algorithm for sparse online learning}
Given Data Input~\ref{data_input}, the following computation can be performed in superposition on indices $j\in[d]$, which allows us to efficiently compute each entry of the weight vector $w^{(t)}$ at any time step $t$. We defer the proof to Appendix~\ref{app:proof_big_CD}. Similar unitaries were studied in, e.g., ~\cite{chakrabarti2021threshold, li2019sublinear, rebentrost2021quantum, vedral1996quantum}. 
\begin{lemma}\label{lem:big_GD}
Let $\theta\in\mathbb R_{>0}$. For all $t\in[T]$, given example oracles $U_{x^{(t)}}$ as in Data Input~\ref{data_input}, vectors $y = (y^{(1)}, \cdots, y^{(t)}), \tilde y = (\tilde y^{(1)}, \cdots, \tilde y^{(t)})\in\mathbb R^{t}$ and a real number $\eta\in\mathbb R_{>0}$. Assuming access to a gravity sequence $(g^{(1)}, \cdots, g^{(T)})$ and a truncation oracle as in Lemma~\ref{lem:truncation}, there exists a unitary operators that perform the operation $\ket{j}\ket{\bar 0} \rightarrow \ket{j}\ket{w^{(t)}_j}$ to sufficient numerical precision, where
\begin{enumerate}[(i)]
    \item \label{lem:GD_unitary_logistic} For logistic regression, $w^{(t)}_j = \mathcal T\left(w_j^{(t-1)} + 2\eta \frac{x_j^{(t)}y^{(t)}e^{-y^{(t)}\tilde y^{(t)}}}{1 + e^{-y^{(t)}\tilde y^{(t)}}}, g^{(t)}\eta, \theta\right)$;
    \item \label{lem:GD_unitary_SVM} For the SVM, $w^{(t)}_j = \begin{cases}\mathcal T\left(w^{(t)}_j + \eta y^{(t)}x_j^{(t)}, g^{(t)}\eta, \theta\right), \text{ if } y^{(t)}\tilde y^{(t)} < 1\\
        \mathcal T\left(w^{(t)}_j, g^{(t)}\eta, \theta\right), \text{ otherwise } 
    \end{cases}$;
    \item \label{lem:GD_unitary} For least squares, $w^{(t)}_j = \mathcal T\left(w^{(t)}_j + 2\eta \left(y^{(t)} - \tilde y^{(t)}\right)x^{(t)}_j, g^{(t)}\eta, \theta\right)$. 
\end{enumerate}
This computation takes $O(T)$ queries to the data input and requires $O(T +\log d)$ qubits and quantum gates. 
\end{lemma}

We present our quantum algorithm for sparse online learning in Algorithm~\ref{algo:QSOL_main}. Note that this quantum algorithm applies to logistic regression, the SVM and least squares when choosing their respective unitaries from Lemma~\ref{lem:big_GD} in Line 5. 
\begin{algorithm}[H]
\caption{Quantum algorithm for sparse online learning with truncated gradient descent}
\label{algo:QSOL_main}
\begin{algorithmic}[1]
\REQUIRE Threshold $\theta>0$, gravity sequence $\{g^{(1)},\cdots, g^{(T)}\}\leq g_{\max}$, learning rate $\eta\in (0, 1)$, $\tilde y^{(1)} = 0$, failure probability $\delta$, errors $\epsilon_{\textup{IP}}, \epsilon_{\textup{norm}}\in (0, 1)$.
\FOR {$t=1$ to $T$}
\STATE Receive example oracle $U_{x^{(t)}}$.  
\STATE Compute the estimate $\tilde y^{(t)}$ of the inner product $\hat y^{(t)} = \sum_{j=1}^d w^{(t)}_j x^{(t)}_j$ up to additive accuracy $\epsilon_{\text{IP}}$ with success probability $1-\frac{\delta}{3T}$ using Lemma~\ref{lem:quantum_inner_product_estimation}. 
\STATE Receive the true label $y^{(t)}$. 
\STATE Prepare the state $\ket{w^{(t+1)}}$ with success probability $1-\frac{\delta}{3T}$ using Lemma~\ref{lem:truncation}, ~\ref{lem:quantum_norm_est_and_quantum_state_prep}(\ref{lem:quantum_state_preparation}) and Lemma~\ref{lem:big_GD} (depending on the problem). 
\STATE Obtain an estimate $\tilde q^{(t+1)}$ of  $\lVert w^{(t)}\cdot I\left(\left\vert w^{(t)}\right\vert \leq \theta\right)\rVert_1$ using Lemma~\ref{lem:quantum_norm_est_and_quantum_state_prep}(\ref{lem:quantum_norm_estimation_multiplicative}) up to additive error $\epsilon_{\text{norm}}$ with success probability $1-\frac{\delta}{3T}$. 
\ENDFOR
\end{algorithmic}
\end{algorithm}

We emphasize that each entry of the weight vector can be efficiently computed, independently of the other entries as shown in Lemma~\ref{lem:big_GD}. This includes the truncation step. Hence, it follows that the weight vectors need not be stored in qubits, thereby saving on the space/memory of the algorithm. For the inner product estimation, we access the entries of the weight vector by simply computing them, thanks to efficient unitaries for arithmetic computation and truncation. These entries, after the truncation, will not be entangled with auxiliary qubits.

\section{Convergence and Time Complexity Analysis} \label{sec:analysis}
In this subsection, we analyze the regret and time complexity of Algorithm~\ref{algo:QSOL_main} applied to logistic regression,  SVM, and least squares, respectively. We show that our quantum algorithms achieve a quadratic speedup in the data dimension $d$ over classical algorithms.

\subsection{Quantum sparse online algorithm for logistic regression}\label{Quantum algorithm:LR}
The quantum sparse online algorithm for logistic regression is obtained by applying Algorithm~\ref{algo:QSOL_main} with Lemmas~\ref{lem:big_GD}(\ref{lem:GD_unitary_logistic}). 
The regret is guaranteed by Theorem~\ref{lem:regret}, with its proof deferred to Appendix~\ref{app:proof_regret}.
\begin{theorem}[Regret for online logistic regression]\label{lem:regret}
Let $\delta \in(0, 1)$. For all $t\in[T]$, let $\tilde y^{(t)}$ be an estimate of $\hat y^{(t)} = w^{(t)T}x^{(t)}$ to additive error $\epsilon_{\textup{IP}}$, and  $\tilde q^{(t+1)}$ be an estimate of $q^{(t+1)} := \Vert w^{(t+1)}\cdot I(\vert w^{(t+1)}\vert\leq \theta)\Vert_1$ to additive error $\epsilon_{\textup{norm}}$. Set $\eta = \frac{1}{C^2\sqrt{T}}$,  $\epsilon_{\textup{norm}} = \frac{1}{2\eta T}$ and $\epsilon_{\textup{IP}} = \frac{1}{2\sqrt T}$. Under Assumption~\ref{ass: regret_ass}(\ref{ass:sup_x}), Algorithm~\ref{algo:QSOL_main} with Lemma~\ref{lem:big_GD}(\ref{lem:GD_unitary_logistic})  achieves a regret bound of 
\be\nonumber
& & \frac{1}{T}\sum_{t=1}^T \ln \left(1 + e^{-y^{(t)}\tilde y^{(t)}}\right) + \frac{1}{T}\sum_{t=1}^T g^{(t)}\tilde q^{(t)} -\frac{1}{T}\sum_{t=1}^T \ln \left(1 + e^{-y^{(t)}u^T x^{(t)}}\right) \\\nonumber
& &- \frac{1}{T}\sum_{t=1}^T g^{(t)}\left\Vert u\cdot I\left(\left\vert w^{(t+1)}\right\vert\right)\right\Vert_1\leq \frac{1 + C^2\left(2 + g_{\max} + \Vert u^*\Vert^2_2\right)}{2\sqrt T}\\\nonumber
\ee
with success probability $1-\delta$ using $O\left(T^{5/2}\sqrt{d}\log\frac{T}{\delta}\right)$ queries, where $g_{\max} = \max_{t \in [T]} g^{(t)}$.
\end{theorem}

By taking $T = \Theta(C^4\Vert u^*\Vert_2^4/\epsilon^2)$, the regret in Theorem~\ref{lem:regret} becomes $\Theta(\epsilon)$.
This implies a quantum algorithm for (offline) logistic regression with time complexity $\tilde O(C^{10} \Vert u^*\Vert_2^{10}\sqrt d /\epsilon^{5})$, which achieves an exponential improvement in the dependence on $C\Vert u^*\Vert_2$ compared to the prior best offline result $\tilde O(C^2\Vert u^*\Vert_2 \exp(C\Vert u^*\Vert_2)/\epsilon^2)$ due to \cite{shao2019fast}.

\subsection{Quantum sparse online algorithm for SVM}\label{Quantum algorithm:SVM}
The quantum sparse online algorithm for SVM is obtained by applying Algorithm~\ref{algo:QSOL_main} with Lemma~\ref{lem:big_GD}(\ref{lem:GD_unitary_SVM}). 
The regret is guaranteed by Theorem~\ref{theorem:regret_SVM}, with its proof deferred to Appendix~\ref{app:proof_regret_SVM}.
\begin{theorem}[Regret for online hinge loss]\label{theorem:regret_SVM}
Let $\delta \in(0, 1)$. For all $t\in[T]$, let $\tilde y^{(t)}$ be an estimate of $\hat y^{(t)} = w^{(t)T}x^{(t)}$ to additive error $\epsilon_{\textup{IP}}$ and $\tilde q^{(t+1)}$ be an estimate of $q^{(t+1)} := \left\Vert w^{(t+1)}\cdot I\left(\left\vert w^{(t+1)}\right\vert\leq \theta\right)\right\Vert_1$ to additive error $\epsilon_{\textup{norm}}$. Set $\eta = \frac{1}{C^2 T^2}$ and $\epsilon_{\textup{IP}} = \epsilon_{\textup{norm}} = \frac{1}{2\sqrt T}$. Under Assumption~\ref{ass: regret_ass}(\ref{ass:sup_x}), Algorithm \ref{algo:QSOL_main} with Lemma~\ref{lem:big_GD}(\ref{lem:GD_unitary_SVM}) achieves a regret bound of 
\be\nonumber
& & \frac{1}{T}\sum_{t=1}^T \left(1-y^{(t)}\tilde y^{(t)}\right)^+ + \frac{1}{T}\sum_{t=1}^T g^{(t)}\tilde q^{(t+1)} -\frac{1}{T}\sum_{t=1}^T \left(1 -y^{(t)}u^T x^{(t)}\right)^+ \\\nonumber
& & - \frac{1}{T}\sum_{t=1}^T g^{(t)}\left\Vert u\cdot I\left(\left\vert w^{(t+1)}\right\vert\right)\right\Vert_1 \leq \frac{2 + C^2\left(g_{\max} + \Vert u^*\Vert_2^2\right)}{2\sqrt T}
\ee
with success probability $1-\delta$ using $O\left(T^{5/2}\sqrt{d}\log\frac{T}{\delta}\right)$ queries, where $g_{\max} = \max_{t \in [T]} g^{(t)}$.
\end{theorem}

By taking $T = \Theta(C^4\Vert u^*\Vert_2^4/\epsilon^2)$, the regret in Theorem~\ref{theorem:regret_SVM} becomes $\Theta(\epsilon)$.
This implies a quantum algorithm for (offline) SVM with time complexity $\tilde O(C^{10} \Vert u^*\Vert_2^{10}\sqrt d /\epsilon^{5})$, achieving a better dependence on $\epsilon$ than the prior best offline result $\tilde O(\sqrt d C^2 \Vert u^*\Vert_2^2/\epsilon^5 + \sqrt d/\epsilon^8)$ due to \cite{li2019sublinear}.

\subsection{Quantum sparse online algorithm for least squares}\label{Quantum algorithm:LS}

The quantum sparse online algorithm for least squares is obtained by applying Algorithm~\ref{algo:QSOL_main} with Lemma~\ref{lem:big_GD}(\ref{lem:GD_unitary}).
It turns out that the quantum speedup appears when the prediction error is constant-bounded. 
Here, the prediction error means the distance between true and predicted labels, which was also considered previously in the literature, e.g., \cite{lin2022bounded}.
We formally state this condition as follows. 
\begin{assumption}[Constant-bounded prediction error]\label{ass:bound_D} 
    For any $t\in[T]$, let $y^{(t)}$ be the true label and let $\hat y^{(t)} = \sum_{j=1}^d w_j^{(t)}\cdot x_j^{(t)}$ be the predicted label. The prediction error is bounded by a constant $D\in\mathbb R_+$, such that for all $t$, $\left\vert y^{(t)} - \hat y^{(t)}\right\vert\leq D$. 
\end{assumption}

Under Assumption~\ref{ass:bound_D}, the regret is guaranteed by Theorem~\ref{thm: QSOL_LS}, with its proof in Appendix~\ref{app:proof_QSOL_LS}.

\begin{theorem}[Regret for online least squares]\label{thm: QSOL_LS}
Let $\delta\in(0, 1)$. Let $u\in\mathbb R^d$ be any vector and for $t\in[T]$, let $\tilde y^{(t)}$ be an estimate of $\hat y^{(t)} = w^{(t)T}x^{(t)}$ to additive error $\epsilon_{\textup{IP}}$ and $\tilde q^{(t+1)}$ be an estimate of $q^{(t+1)} := \left\Vert w^{(t+1)}\cdot I\left(\left\vert w^{(t+1)}\right\vert\leq \theta\right)\right\Vert_1$ to additive error $\epsilon_{\textup{norm}}$. Set $\eta = \frac{1}{C^2\sqrt{T}}$, $\epsilon_{\textup{IP}} = \frac{1}{2\sqrt T}$ and $\epsilon_{\text{norm}} = \frac{1}{2\eta T}$. Under Assumptions~\ref{ass: regret_ass}(\ref{ass:sup_x}) and~\ref{ass:bound_D}, Algorithm~\ref{algo:QSOL_main} with Lemma~\ref{lem:big_GD}(\ref{lem:GD_unitary}) achieves a regret bound of
\be
& & \frac{1}{T}\sum_{t=1}^T\left(\tilde y^{(t)} - y^{(t)}\right)^2 + \frac{1}{T}\sum_{t=1}^T g^{(t)}\tilde q^{(t+1)}  - \frac{1}{T}\sum_{t=1}^T \left(u^T x^{(t)} - y^{(t)}\right)^2
\nonumber \\  \nonumber
& &  - \frac{1}{T}\sum_{t=1}^T g^{(t)}\left\Vert u\cdot I\left(\left\vert w^{(t+1)}\right\vert\leq\theta\right)\right\Vert_1
\leq \frac{C^2\left(CD + g_{\max} + \Vert u^*\Vert^2_2\right)}{\sqrt T} 
\ee
with probability at least $1-\delta$ using $O\left(T^{5/2}\sqrt{d}\log\frac{T}{\delta}\right)$ queries, where $g_{\max} = \max_{t \in [T]} g^{(t)}$. 
\end{theorem}

By setting $T = \Theta((C^6 + C^4 \Vert u^*\Vert_2^4)/\epsilon^2)$, the regret in Theorem~\ref{thm: QSOL_LS} becomes $\Theta(\epsilon)$.
This implies a quantum algorithm for (offline) least squares with time complexity $\tilde O((C^{15} + C^{10} \Vert u^*\Vert_2^{10})\sqrt d /\epsilon^{5})$.
For comparison, we are aware of a quantum algorithm for offline least squares proposed in \cite{liu2017fast} that considers different conditions and parameters.
Their algorithm has time complexity $\tilde O(s^2\kappa^2/\epsilon^2)$, where $s$ denotes the sparsity of the data matrix and $\kappa$ is its condition number. 

\section{Discussion and Conclusion}\label{sec:conclusion}
We propose a quantum online learning algorithm that outputs sparse solutions. Our quantum algorithm can be applied to logistic regression, the SVM and least squares. We show that the quantum algorithm achieves a quadratic speedup in the dimension of the problem as compared to its classical counterpart. The speedup stems from the use of quantum subroutines based on quantum amplitude estimation and amplification. We note that the speedup is only noticeable when $d\geq \Omega(T^5 \log^2 (T/\delta))$, which makes the algorithm useful for high-dimensional learning tasks. As our quantum algorithm is erroneous, it is natural that convergence is achieved after a greater number of steps as compared to its classical analogue.  Our algorithm maintains a regret bound of the same order as compared to the classical algorithm of \cite{langford2009sparse}, i.e. $O(1/\sqrt T)$. We leverage unitaries that perform arithmetic computations, which allows us to save on the space/memory of the algorithm for storing the weight vector, which is $O(d)$ in \cite{langford2009sparse}. 

Despite our quantum algorithm having a running time that achieves quadratic improvement in the dimension $d$ of the weight vector, its dependence on the number of time steps $T$ increases. One natural question would be to ask if the trade-off between $T$ and $d$ can be avoided. Besides that, it would be interesting to explore how other variants of gradient descent such as mirror descent or stochastic gradient descent, combined with different ``feature selection" techniques to obtain sparse solutions can contribute to an improvement in the regret bound. Considering that we have a unitary that computes entries of the weight vector that is updated via truncated gradient descent, one could consider potential applications of this unitary, for example in reinforcement learning~\cite{mahadevan2012sparse}. On the other hand, one could explore possible applications of quantum algorithms in obtaining sparse solutions in the online learning setting as there has not been any work done in this regime. Instead of analyzing the (static) regret, one could consider studying the \textit{dynamic} regret of the online algorithm which can be useful in scenarios where the optimal solution keeps changing in evolving environments~\cite{besbes2015non, jadbabaie2015online, mokhtari2016online, yang2016tracking, zhang2017improved, zhang2018dynamic, zhao2020dynamic}. 

\section*{Acknowledgements}
The authors acknowledge helpful discussions with Naixu Guo and Zhan Yu. 

The work of Debbie Lim, Yixian Qiu, and Patrick Rebentrost was supported by the National Research Foundation, Singapore and A*STAR under its CQT Bridging Grant and its Quantum Engineering Programme under grant \mbox{NRF2021-QEP2-02-P05}. The work of Debbie Lim was also supported in part by the QuantERA Project QOPT and the Latvian Quantum Initiative under EU Recovery and Resilience Facility under project no.\ \mbox{2.3.1.1.i.0/1/22/I/CFLA/001}.
The work of Qisheng Wang was supported by the Engineering and Physical Sciences Research Council under Grant \mbox{EP/X026167/1}.

\addcontentsline{toc}{section}{References}
    
\bibliographystyle{alphaurl}
\bibliography{iclr2025_conference}

\appendix

\section{Algorithms in Prior Work}\label{app:algo}
The easiest algorithm for addressing unconstrained smooth optimization problems is gradient descent. For a convex and differentiable function $L:\mathbb R^d\rightarrow\mathbb R^d$, gradient descent iteratively updates the solution by moving in the direction of the negative gradient to minimize the objective function in unconstrained smooth optimization problems.   
\begin{algorithm}[H]
\caption{Gradient descent~\cite{hazan2016introduction}}
\label{algo:GD}
\begin{algorithmic}[1]
\REQUIRE Loss function $L$, total time steps $T$, initial point $w^{(1)}$, learning rates $\eta^{(1)},\cdots, \eta^{(T)}$
\ENSURE $w^{(T+1)}$
\FOR {$t=1$ \TO $T$}
\STATE Let $w^{(t+1)} = w^{(t)} - \eta^{(t)}\nabla L(w^{(t)})$.
\ENDFOR
\end{algorithmic}
\end{algorithm}

Gradient descent is modified to truncate the solution after $K$ steps to maintain its sparsity. The truncation is performed on entries of the solution that fall within a certain threshold value. This truncated solution is then used in the next iteration of gradient descent.    
\begin{breakablealgorithm}
\caption{Truncated gradient descent~\cite{langford2009sparse}}
\label{algo:TGD}
\begin{algorithmic}[1]
\REQUIRE Convex loss function $L$, total time steps $T$, initial point $w^{(1)}$, learning rates $\eta^{(1)},\cdots, \eta^{(T)}$, threshold $\theta$, $K$, shrinkage parameters $\alpha^{(1)}, \cdots , \alpha^{(T)}$. 
\FOR {$t=1$ \TO $T$}
\STATE Let $w^{\prime(t+1)} = w^{(t)} + \eta^{(t)}\nabla L(w^{(t)})$. 
\FOR {$j=1, \cdots, d$}
\IF {$0\leq w^{\prime(t+1)}\leq \theta$ and $\frac{t}{K}$ is an integer}
\STATE $w^{(t+1)}_j = \max\left\{w^{\prime(t+1)}_j - \alpha^{(t)}, 0\right\}$.
\ELSIF {$-\theta\leq w^{\prime(t+1)}\leq 0$ and $\frac{t}{K}$ is an integer}
\STATE $w^{(t+1)}_j = \min\left\{w^{\prime(t+1)}_j + \alpha^{(t)}, 0\right\}$.
\ELSE
\STATE $w^{(t+1)}_j = w^{\prime(t+1)}_j$.
\ENDIF
\ENDFOR
\ENDFOR
\ENSURE $w^{(T+1)}$. 
\end{algorithmic}
\end{breakablealgorithm}

Truncated gradient descent is used in the online learning framework to output sparse solutions. In every iteration $t\in[T]$, the algorithm receives an unlabelled example $x^{(t)}$, makes a prediction $\hat y^{(t)}$ on the label using the unlabelled example and the weight vector computed from the previous iteration. After that, it receives the true label $y^{(t)}$. The algorithm then updates its weight vector using truncated gradient descent.  

\begin{breakablealgorithm}
\caption{Online sparse learning algorithm with truncated gradient descent~\cite{langford2009sparse} }
\label{algo}
\begin{algorithmic}[1]
\REQUIRE Threshold $\theta>0$, gravity sequence $\{g^{(1)},\cdots, g^{(T)}\}\leq g_{\max}$, learning rate $\eta\in (0, 1)$, example oracle $\mathcal O$. 
\STATE Initialize $w^{(1)}=\left (0, \cdots, 0\right)\in\mathbb{R}^d$.
\FOR {$t=1$ to $T$}
\STATE Receive unlabeled example $x^{(t)} = \left(x^{(t)}_1, \cdots, x^{(t)}_d\right)\in\mathbb R^d$ form example oracle $\mathcal O$. 
\STATE Compute the prediction $\hat y^{(t)} = \sum_{j=1}^d w^{(t)}_j x^{(t)}_j$.
\STATE Receive the true label $y^{(t)}$ from example oracle $\mathcal O$. 
\FOR {$j=1$ to $d$}
\STATE $w^{\prime(t+1)}_j\gets w^{(t)}_j - \eta\nabla L\left(w^{(t)}\right)$.
\ENDFOR
\FOR {$j=1$ to $d$}
\IF {$0\leq w^{(t)}_j\leq \theta$}
\STATE $w^{(t)}_j\gets \max\left\{w^{\prime(t)}_j - g^{(t)}\eta, 0\right\}$
\ELSIF {$-\theta\leq w^{(t)}_j\leq 0$}
\STATE $w^{(t)}_j\gets \min\left\{w^{\prime(t)}_j+g^{(t)}\eta, 0\right\}$
\ELSE 
\STATE $w^{(t)}_j\gets w^{(t)}_j$
\ENDIF
\ENDFOR
\ENDFOR
\end{algorithmic}
\end{breakablealgorithm}

\section{Omitted Proofs in Section~\ref{sec:aquantum_algorithm}}

\subsection{Proof of Lemma~\ref{lem:truncation}}\label{app:truncation_unitary}
    Start with the state $\ket{\alpha} \ket{-\theta}\ket{0}\ket{\theta}\ket{x}\ket{0}\ket{0}\ket{0}\ket{0}$. Query the Between oracle on the third to sixth registers. Then, 
    \begin{enumerate}[(i)]
        \item Condition on the sixth register being $\ket{1}$, flip the seventh register to flag that $0\leq x\leq \theta$, and do
        \begin{equation*}
            \ket{\alpha} \ket{-\theta}\ket{0}\ket{\theta}\ket{x}\ket{1}\ket{1}\ket{0}\ket{0} \rightarrow \ket{\alpha} \ket{-\theta}\ket{0}\ket{\theta}\ket{x}\ket{1}\ket{1}\ket{x - \alpha}\ket{0}. 
        \end{equation*}
        Next, query the Minmax oracle on the seventh and eighth register to get 
        \begin{equation*}
           \ket{\alpha} \ket{-\theta}\ket{0}\ket{\theta}\ket{x}\ket{1}\ket{1}\ket{x - \alpha}\ket{0}\rightarrow \ket{\alpha} \ket{-\theta}\ket{0}\ket{\theta}\ket{x}\ket{1}\ket{1}\ket{x - \alpha}\ket{\max(x - \alpha, 0)}.
        \end{equation*}
        Lastly, swap the fifth and the last registers and uncompute intermediate registers.
    \item Condition on the sixth register being $\ket{0}$, query the between oracle on the second, third, fifth and sixth registers. Then, 
    \begin{enumerate}[(i)]
        \item condition on the sixth being $\ket{1}$, do
        \begin{equation*}
            \ket{\alpha} \ket{-\theta}\ket{0}\ket{\theta}\ket{x}\ket{1}\ket{0}\ket{0}\ket{0} \rightarrow \ket{\alpha} \ket{-\theta}\ket{0}\ket{\theta}\ket{x}\ket{1}\ket{0}\ket{x + \alpha}\ket{0}. 
        \end{equation*}
        Next, query the Minmax oracle on the eighth and ninth registers to get 
        \begin{equation*}
            \ket{\alpha} \ket{-\theta}\ket{0}\ket{\theta}\ket{x}\ket{1}\ket{0}\ket{x + \alpha}\ket{0} \rightarrow \ket{\alpha} \ket{-\theta}\ket{0}\ket{\theta}\ket{x}\ket{1}\ket{0}\ket{x + \alpha}\ket{\min(x + \alpha, 0)}.
        \end{equation*}
        Lastly, swap the sixth and the last registers and uncompute intermediate registers.
        \item Condition on the sixth register being $\ket{0}$, do nothing. 
    \end{enumerate}
    Since we assume the use of the quantum arithmetic model, we hence obtain a running time of $\tilde O(1)$. 
    \end{enumerate} 

\subsection{Proof of Lemma~\ref{lem:quantum_norm_est_and_quantum_state_prep}}\label{app:proof_quantum_norm_est_and_quantum_state_prep}

\begin{enumerate}[(i)]
\item Using the query access, create the circuit to prepare the state $\frac{1}{\sqrt{d}}\sum_{j=1}^d \ket{j}\ket{u_j}\ket{0}$. Use quantum maximum finding~\cite{durr1996quantum} to find 
\begin{equation*}
    \lVert u\rVert_{\infty}:= \max_{j\in[d]}\vert u_j \vert
\end{equation*}
with success probability $1-\delta/2$. Apply a controlled-rotation to the state obtains 
\be\label{state:controlled-rotation}
\frac{1}{\sqrt{d}}\sum_{j=1}^d \ket{j}\ket{u_j}\left(\sqrt{\frac{\vert u_j\vert}{\lVert u\rVert_{\infty}}}\ket{0} + \sqrt{1 - \frac{\vert u_j\vert}{\lVert u\rVert_{\infty}}}\ket{1}\right).
\ee
Let $U_u$ be the unitary that prepares the state in Eq.~(\ref{state:controlled-rotation}). Define new unitaries $U = U_u(I - 2\ket{\bar 0}\bra{\bar 0})U_u^\dagger$ and $V = I - I\otimes \ket{ 0}\bra{ 0}$. Fact~\ref{fact:new_AE} allows us to obtain an estimate $\tilde a$ of $a = \frac{\lVert u\rVert_1}{d\Vert u\Vert_\infty}$ such that $\vert \mathbb E[\tilde a]-a\vert\leq \frac{\epsilon_0^2}{32}a^2$ 
and $\operatorname{Var}(\tilde a)\leq \frac{91a}{K^2} + \frac{\epsilon_0^2}{32}a^2$, restoring the initial state with success probability at least $1 - \frac{\epsilon_0^2}{32}a^2$, using $O\left(K\log\log K\log(K/\epsilon_0)\right)$ expected number of applications of $U$ and $V$. Setting $K > \frac{8}{\epsilon_0}\sqrt{\frac{91}{a}}$ via exponential search without knowledge of $a$, by Chebyshev's inequality we have 
\begin{align*}
\mathbb P\left[\lvert \tilde a - \mathbb E[\tilde a]\rvert\geq \frac{\epsilon_0}{2}a\right]\leq \frac{4}{\epsilon_0^2a^2} \left(\frac{91 a}{K^2} + \frac{\epsilon_0^2 a^2}{32} \right)\leq \frac{4}{\epsilon_0^2 a^2} \left(\frac{\epsilon_0^2 a^2}{64} + \frac{\epsilon_0^2 a^2}{32} \right)\leq \frac{1}{16} + \frac{1}{8} \leq \frac{1}{4}.
\end{align*}
The success probability $3/4$ is boosted with $O(\log\frac{1}{\delta})$ repetitions to $1-\delta/2$ via the median of means technique. Hence,
\begin{align*}
\lvert \tilde a-a\vert\leq \lvert \tilde a- \mathbb E[\tilde a]\rvert + \lvert \mathbb E[\tilde a] - a\rvert \leq \epsilon_0 a/2 + \epsilon_0 a/2 = \epsilon_0 a
\end{align*}
with success probability at least $1 - 4\delta$.
Hence the quantity $\tilde \Gamma : = \tilde a$ is an estimate 
\begin{align*}
\left\vert \tilde \Gamma - \frac{\Vert u\Vert_1}{d\lVert u\Vert_\infty} \right\vert = \lvert \tilde a - a\rvert  \leq \epsilon_0 a.  
\end{align*}
This brings the total time complexity to \begin{align*}
    O\left(\frac{1}{\epsilon_0}\sqrt{\frac{ d \Vert u\Vert_\infty}{\Vert u\Vert_1}}\log \log\left(\frac{1}{\epsilon_0}\sqrt{\frac{ d \Vert u\Vert_\infty}{\Vert u\Vert_1}}\right) \log \left(\frac{1}{\epsilon_0^2}\left(\frac{ d \Vert u\Vert_\infty}{\Vert u\Vert_1}\right)^{3/2}\right)\log \frac{1}{\delta}\right)
\end{align*} 
in expectation. While \cite{cornelissen2023sublinear} proves a result in expected time, we use the probabilistic result obtained from Markov’s inequality and repetition at a cost of another factor of $O\left(\log\frac{1}{\delta}\right)$. 

We note that the additive version of norm estimation can be easily specialized by setting $K > \frac{8}{\epsilon_0}\sqrt{91 a}$ and has the same time and gate complexity. 
\item Note that the state in Eq.~(\ref{state:controlled-rotation}) can be rewritten as 
\begin{equation*}
    \sqrt{\frac{\lVert u\rVert_1}{\lVert u\rVert_{\infty}d}}\sum_{j=1}^d \ket{j}\ket{u_j}\left(\sqrt{\frac{\vert u_j\vert}{\lVert u\rVert_1}}\ket{0} + \sqrt{1 - \frac{\vert u_j \vert}{\lVert u\rVert_1}}\ket{1}\right). 
\end{equation*}
Amplify the $\ket{0}$ part via amplitude amplification~\cite{brassard2002quantum} to obtain $\vert u\rangle = \sum_{j=1}^d \sqrt{\frac{\vert u_j\vert}{\lVert u\rVert_1}}\ket{j}$ with success probability $1-\delta$ using $O\left(\sqrt{\frac{\lVert u\rVert_\infty d}{\lVert u\rVert_1}}\log\frac{1}{\delta}\right) \subseteq O\left(\sqrt{ d}\log\frac{1}{\delta}\right)$ calls to $U_u$ and $\tilde O\left(\sqrt{ d} \log\frac{1}{\zeta}\log\frac{1}{\delta}\right)$ gates. For all $j\in[d]$, we have $\tilde p_j = \frac{u_j}{\tilde \Gamma}$. Also, note that $\lVert u\rVert_1 - \tilde \Gamma \leq \vert \lVert u\rVert_1 - \tilde \Gamma \vert \leq \zeta\lVert u\rVert_1$, and hence $\frac{1}{\tilde \Gamma}\geq \frac{1}{\lVert u\rVert_1(1-\zeta)}$. Therefore, 
\begin{align*}
    \left\lVert \tilde p - \frac{u}{\lVert u\rVert_1}\right\rVert_1 
    = \sum_{j=1}^d \left\vert\frac{u_j}{\tilde\Gamma} - \frac{u_j}{\lVert u\rVert_1}\right\vert 
    \leq \sum_{j=1}^d \left\vert \frac{u_j\cdot\zeta\lVert u\rVert_1}{\lVert u\rVert_1 \cdot (1-\zeta)\lVert u\rVert_1}\right\vert
    = \frac{\zeta}{(1-\zeta)}\sum_{j=1}^d \left\vert \frac{u_j}{\lVert u\rVert_1}\right\vert 
    = \frac{\zeta}{(1-\zeta)}\leq 2\zeta.
\end{align*}
\end{enumerate}

\subsection{Proof of Lemma~\ref{lem:quantum_inner_product_estimation}}\label{app_proof_quantum_inner_product_estimation}
Define vectors $u^+$ and $u^-$ as follows
\begin{equation*}
    u^+_i  = 
\begin{cases}
    u_i, \text{ if } \text{sign}(u_i) = 1\\
    0, \text{ otherwise}\\
\end{cases}, 
\hspace{3cm}
u^-_i  = 
\begin{cases}
    0, \text{ if } \text{sign}(u_i) = 1\\
    -u_i, \text{ otherwise}\\
\end{cases}
\end{equation*}
Notice that $u = u^+ - u^-$. Define $v^+$ and $v^-$ in a similar way. Then
\begin{equation*}
u\cdot v = u^+\cdot v^+ + u^-\cdot v^-  - u^-\cdot v^+ - u^+\cdot v^-
\end{equation*}
Let $z^+$ and $z^-$ be vectors such that $z^+_i = u^+_i \cdot v^+_i + u^-_i \cdot v^-_i $ and $z^-_i = u^+_i \cdot v^-_i + u^-_i \cdot v^+_i$. Then, observe that 
\begin{equation*}
u\cdot v = \lVert z^+\rVert_1 - \lVert z^-\rVert_1
\end{equation*}
Next, determine $z^+_{\max} := \max_{j\in[d]} \vert z^+_j\vert$ using quantum maximum finding with success probability $1-\delta/4$ using $O\left(\sqrt{d}\log\frac{1}{\delta}\right)$ queries and $\tilde O\left(\sqrt{d}\log\frac{1}{\delta}\right)$ quantum gates~\cite{durr1996quantum}. If $z^+_{\max} = 0$ up to sufficient numerical accuracy, then $\widetilde{IP} =0 $ and we are done. Otherwise, use Lemma~\ref{lem:quantum_norm_est_and_quantum_state_prep}(\ref{lem:quantum_norm_estimation_multiplicative}) to obtain an estimate $\tilde\Gamma_{z^+}$  of $\left\Vert \frac{z^+}{z^+_{\max}}\right\Vert_1$ such that 
\begin{equation*}
\left\vert \left\Vert \frac{z^+}{z^+_{\max}}\right\Vert_1  - \tilde\Gamma_{z^+}\right\vert\leq \frac{\epsilon^+}{2}\left\Vert \frac{z^+}{z^+_{\max}}\right\Vert_1
\end{equation*}
with success probability at least $1-\delta/4$ with $\tilde O\left(\frac{1}{\epsilon^+}\sqrt{\frac{dz^+_{\max}}{\lVert z^+ \rVert_1}}\log \frac{1}{\delta}\right)\subseteq \tilde O\left(\frac{\sqrt{d}}{\epsilon^+}\log \frac{1}{\delta}\right)$ queries and $\tilde O\left(\frac{\sqrt{d}}{\epsilon^+}\log \frac{1}{\delta}\right)$ quantum gates.

Similarly for the case of $z^-$, find $z^-_{\max} := \max_{j\in[d]} \vert z^-\vert$ using quantum maximum finding with success probability $1-\delta/4$ using $O\left(\sqrt{d}\log\frac{1}{\delta}\right)$ queries and $\tilde O\left(\sqrt{d}\log\frac{1}{\delta}\right)$ quantum gates~\cite{durr1996quantum}. If $z^-_{\max} = 0$ up to sufficient numerical accuracy, then $\widetilde{IP} =0 $ and we are done. Otherwise, use Lemma~\ref{lem:quantum_norm_est_and_quantum_state_prep}(\ref{lem:quantum_norm_estimation_multiplicative}) to obtain an estimate $\tilde\Gamma_{z^-}$  of $\left\Vert \frac{z^-}{z^-_{\max}}\right\Vert_1$ such that 
\begin{equation*}
\left\vert \left\Vert \frac{z^-}{z^-_{\max}}\right\Vert_1  - \tilde\Gamma_{z^-}\right\vert\leq \frac{\epsilon^-}{2}\left\Vert \frac{z^-}{z^-_{\max}}\right\Vert_1 
\end{equation*}
with success probability at least $1-\delta/4$ with $\tilde O\left(\frac{1}{\epsilon^-}\sqrt{\frac{dz^+_{\max}}{\lVert z^+ \rVert_1}}\log \frac{1}{\delta}\right)\subseteq O\left(\frac{\sqrt{d}}{\epsilon^-}\log \frac{1}{\delta}\right)$ queries and $\tilde O\left(\frac{\sqrt{d}}{\epsilon^-}\log \frac{1}{\delta}\right)$ quantum gates.

Set $\widetilde{IP} = z^+_{\max}\tilde\Gamma_{z^+} - z^-_{\max}\tilde\Gamma_{z^-}$. Then 
\begin{align*}
\left\vert u\cdot v - \widetilde{IP} \right\vert 
& =  \left\vert \lVert z^+\rVert_1 - \lVert z^-\rVert_1 -  \left(z^+_{\max}\tilde\Gamma^+ - z^-_{\max}\tilde\Gamma^-\right)\right\vert \\
\text{(triangle ineq.)} & \leq  \left\vert \lVert z^+\rVert_1 - z^+_{\max}\tilde\Gamma^+\right\vert + \left\vert \lVert z^-\rVert_1 - z^-_{\max}\tilde\Gamma^-\right\vert\\
\text{(Lemma~\ref{lem:quantum_norm_est_and_quantum_state_prep}(\ref{lem:quantum_norm_estimation_multiplicative}))}& \leq  \frac{\epsilon^+}{2}\lVert z^+\rVert_1 + \frac{\epsilon^-}{2}\lVert z^-\rVert_1\\
& =  \frac{\epsilon^+}{2}\left(\Vert u^+\cdot v^+\Vert_1 + \Vert u^-\cdot v^-\Vert_1\right) + \frac{\epsilon^-}{2}\left(u^+\cdot v^- + u^-\cdot v^+\right)\\
\text{(H\"{o}lder's ineq.)} & \leq  \frac{\epsilon^+}{2}\left(\lVert u^+\rVert_\infty \lVert v^+\rVert_1 + \lVert u^-\rVert_\infty \lVert v^-\rVert_1 \right) + \frac{\epsilon^-}{2}\left(\lVert u^+\rVert_\infty\lVert v^-\rVert_1 + \lVert u^-\rVert_\infty \lVert v^+\rVert_1 \right)\\
& \leq  \frac{\epsilon^+}{2}\Vert u\Vert_\infty\left(\Vert v^+\Vert_1 + \Vert v^-\Vert_1\right) + \frac{\epsilon^-}{2}\Vert u\Vert_\infty\left(\Vert v^+\Vert_1 + \Vert v^-\Vert_1\right)\\
& =  \left(\frac{\epsilon^+}{2} + \frac{\epsilon^-}{2}\right)\Vert u\Vert_\infty\Vert v\Vert_1 
\end{align*}
Setting $\epsilon^+ = \epsilon^- = \frac{\epsilon}{\Vert u\Vert_\infty\Vert v\Vert_1 }$ yields the desired result. The total time complexity is $\tilde O\left(\frac{\lVert u\rVert_\infty\lVert v\rVert_1\sqrt{d}}{\epsilon}\log \frac{1}{\delta}\right)$. 

\subsection{Proof of Lemma~\ref{lem:big_GD}}\label{app:proof_big_CD}
With a computational register of $O(T)$ ancilla qubits for the examples, perform  
        \begin{align}
            \nonumber \ket{ j}\ket{ \bar 0} & \rightarrow \ket{ j }\ket{ x^{(1)}_j } \cdots \ket{x_j^{(t)}}\ket{ \bar 0}\\
            \label{eqn:prep} & \rightarrow \ket{ j }\ket{ x^{(1)}_j } \cdots \ket{x_j^{(t)}}\ket{w_j^{\prime(1)}}\ket{\bar 0}
        \end{align}
    \begin{enumerate}[(i)]
        \item  For $t\in[T]$, let $w_j^{\prime(t)} = w_j^{(t-1)} + 2\eta \left(y^{(t-1)} - \tilde y^{(t-1)}\right)x_j^{(t-1)}$ and  $w^{(t)}_j = \mathcal T\left(w_j^{\prime(t)}, g^{(t)}\eta, \theta\right)$. Then, from Eq. (\ref{eqn:prep}), perform the following operations:
        \begin{align*}
            & \xrightarrow{U_{\mathcal T, g^{(1)}\eta, \theta}}  \ket{ j }\ket{ x^{(1)}_j } \cdots \ket{x_j^{(t)}}\ket{w_j^{\prime(1)}}\ket{w_j^{(1)}}\ket{\bar 0}\\
            & \vdots\\
            & \rightarrow \ket{ j }\ket{ x^{(1)}_j } \cdots \ket{x_j^{(t)}}\ket{w_j^{\prime(1)}}\ket{w_j^{(1)}}\cdots \ket{w_j^{\prime(t)}}\ket{\bar 0}\\
            & \xrightarrow{U_{\mathcal T, g^{(t)}\eta, \theta}}  \ket{ j }\ket{ x^{(1)}_j } \cdots \ket{x_j^{(t)}}\ket{w_j^{\prime(1)}}\ket{w_j^{(1)}}\cdots \ket{w_j^{\prime (t)}}\ket{w_j^{(t)}}\\
    \end{align*}
    \item For $t\in[T]$, let $w_j^{\prime(t)} = w_j^{(t-1)} + 2\eta \frac{x_j^{(t)}y^{(t)}e^{-y^{(t)}\tilde y^{(t)}}}{1 + e^{-y^{(t)}\tilde y^{(t)}}}$ and  $w_j^{(t)} = \mathcal T\left(w_j^{\prime(t)}, g^{(t)}\eta, \theta\right)$. Then, from Eq. (\ref{eqn:prep}), perform the following operations:
    \begin{align*}
        & \xrightarrow{U_{\mathcal T, g^{(1)}\eta, \theta}}  \ket{ j }\ket{ x^{(1)}_j } \cdots \ket{x_j^{(t)}}\ket{w_j^{\prime(1)}}\ket{w_j^{(1)}}\ket{\bar 0}\\
        & \vdots\\
        & \rightarrow \ket{ j }\ket{ x^{(1)}_j } \cdots \ket{x_j^{(t)}}\ket{w_j^{\prime(1)}}\ket{w_j^{(1)}}\cdots \ket{w_j^{\prime(t)}}\ket{\bar 0}\\
        & \xrightarrow{U_{\mathcal T, g^{(t)}\eta, \theta}}  \ket{ j }\ket{ x^{(1)}_j } \cdots \ket{x_j^{(t)}}\ket{w_j^{\prime(1)}}\ket{w_j^{(1)}}\cdots \ket{w_j^{\prime (t)}}\ket{w_j^{(t)}}\\
    \end{align*}
    \item  For $t\in[T]$, let 
    \begin{equation*}
    w_j^{\prime(t)} = \begin{cases}
    w_j^{(t-1)} + \eta y^{(t-1)}x_j^{(t-1)}, \text{ if } y^{(t-1)}\tilde y^{(t-1)} < 1\\
     w_j^{(t-1)}, \text{ otherwise}
    \end{cases}
    \end{equation*}
    and $w_j^{(t)} = \mathcal T\left(w_j^{\prime(t)}, g^{(t)}\eta, \theta\right)$. Then, from Eq. (\ref{eqn:prep}), perform the following operations:
    \begin{align*}
    \ket{ j}\ket{ \bar 0}
    & \rightarrow \ket{ j }\ket{ x^{(1)}_j } \cdots \ket{x_j^{(t)}}\ket{ \bar 0}\\
    & \rightarrow \ket{ j }\ket{ x^{(1)}_j } \cdots \ket{x_j^{(t)}}\ket{w_j^{\prime(1)}}\ket{\bar 0}\\
    & \xrightarrow{U_{\mathcal T, g^{(1)}\eta, \theta}}  \ket{ j }\ket{ x^{(1)}_j } \cdots \ket{x_j^{(t)}}\ket{w_j^{\prime(1)}}\ket{w_j^{(1)}}\ket{\bar 0}\\
    & \vdots\\
    & \rightarrow \ket{ j }\ket{ x^{(1)}_j } \cdots \ket{x_j^{(t)}}\ket{w_j^{\prime(1)}}\ket{w_j^{(1)}}\cdots \ket{w_j^{\prime(t)}}\ket{\bar 0}\\
    & \xrightarrow{U_{\mathcal T, g^{(t)}\eta, \theta}}  \ket{ j }\ket{ x^{(1)}_j } \cdots \ket{x_j^{(t)}}\ket{w_j^{\prime(1)}}\ket{w_j^{(1)}}\cdots \ket{w_j^{\prime (t)}}\ket{w_j^{(t)}}\\
\end{align*}
\end{enumerate}
with sufficient accuracy using the oracles and quantum circuits for arithmetic operations. Uncomputing the intermediate registers will yield the desired result. 

\begin{remark}\label{remArith}
For the addition of two integers each encoded in $k$-bit binary strings, the quantum circuit for addition has a Toffoli count of $2k-1$ and Toffoli depth of $k$~\cite{takahashi2009quantum}. On the other hand, the quantum circuit for integer division has a Toffoli count of $14k^2 +7k + 7$ and a Toffoli depth of $10k + 13$ while subtraction has a Toffoli count of $O(k)$ and a Toffoli depth of $O(1)$~\cite{thapliyal2019quantum}. Circuits for floating-point addition are generated using synthesis tools and can be hand-optimized as shown in ~\cite{haener2018quantum}. Furthermore, circuit sizes for floating-point division have been computed numerically by \cite{gayathri2021t}, which improves upon the work of \cite{nguyen2014resource} and \cite{nachtigal2010design}. In the above computation, we perform $T$ number of divisions and $T$ number of additions, which leads to a circuit of approximately $O(Tk^2)$ size and $O(Tk)$ depth. Since we assume the use of the quantum arithmetic model, we hence obtain the $\tilde O(T+\log n)$ gate complexity. 
\end{remark}

\section{Omitted Proofs in Section~\ref{sec:analysis}}
\subsection{A Fact for Regret Bounds}
To complete our analysis, we need the following regret bounds given in \cite{langford2009sparse}. 
\begin{fact}[{\cite[Theorem 3.1]{langford2009sparse}}]\label{fact:classical_regret}
    Suppose that Assumption~\ref{ass: regret_ass} holds. Then, for all $u\in\mathbb R^d$, Algorithm~\ref{algo} achieves
    \begin{align*}
    & \frac{1 - 0.5A\eta}{T}\sum_{t=1}^T\left[L\left(w^{(t)}, x^{(t)}, y^{(t)}\right) + \frac{g^{(t)}}{1 - 0.5A\eta}\left\Vert w^{(t+1)}\cdot I\left(\left\vert w^{(t+1)}\right\vert\leq \theta\right)\right\Vert_1\right]\\
    & \qquad \leq \frac{\eta}{2}B + \frac{\Vert u\Vert^2_2}{2\eta T} + \frac{1}{T}\sum_{t=1}^T \left[ L\left(u, x^{(t)}, y^{(t)}\right) + g^{(t)} \left\Vert u\cdot I\left(\left\vert w^{(t+1)}\right\vert\leq \theta\right)\right\Vert_1 \right].
    \end{align*}
\end{fact}

\subsection{Proof of Theorem~\ref{lem:regret}}\label{app:proof_regret}
We start by considering the following expression for the regret bound:
\begin{align*}
& \frac{1}{T}\sum_{t=1}^T \ln \left(1 + e^{-y^{(t)}\tilde y^{(t)}}\right) + \frac{1}{T}\sum_{t=1}^T g^{(t)}\tilde q^{(t+1)}\\
& -\frac{1}{T}\sum_{t=1}^T \ln \left(1 + e^{-y^{(t)}u^T x^{(t)}}\right) - \frac{1}{T}\sum_{t=1}^T g^{(t)}\left\Vert u\cdot I\left(\left\vert w^{(t+1)}\right\vert\right)\right\Vert_1.
\end{align*}
Next, we simplify this by separating terms related to $\hat y^{(t)}$ and $\tilde y^{(t)}$ and apply Fact~\ref{fact:classical_regret}:
\begin{align*}
& \leq \frac{1}{T}\sum_{t=1}^T \ln \left(1 + e^{-y^{(t)}\tilde y^{(t)}}\right) + \frac{1}{T}\sum_{t=1}^T g^{(t)}\tilde q^{(t+1)} \\
& \qquad - \frac{1}{T}\sum_{t=1}^T \ln \left(1 + e^{-y^{(t)} \hat y^{(t)}}\right) - \frac{1}{T}\sum_{t=1}^T g^{(t)} q^{(t+1)} + \frac{\eta C^2}{2} + \frac{\lVert u\rVert_2^2}{2\eta T}\\
& \leq \frac{1}{T}\sum_{t=1}^T \left\vert \ln\left(1 + e^{-y^{(t)}\tilde y^{(t)}}\right) - \ln\left(1 + e^{-y^{(t)} \hat y^{(t)}}\right)\right\vert \\
& \qquad + \frac{1}{T}\sum_{t=1}^T g^{(t)}\left\vert \tilde q^{(t+1)} -q^{(t+1)} \right\vert + \frac{\eta C^2}{2} + \frac{\lVert u\rVert_2^2}{2\eta T}.
\end{align*}
Using additive variant of Lemma~\ref{lem:quantum_norm_est_and_quantum_state_prep}(\ref{lem:quantum_norm_estimation_multiplicative}), we further simplify it:
\begin{align*}
& \leq  \frac{1}{T}\sum_{t=1}^T \left\vert \ln\left(1 + e^{-y^{(t)}\tilde y^{(t)}}\right) - \ln\left(1 + e^{-y^{(t)} \hat y^{(t)}}\right)\right\vert  + \frac{\epsilon_{\text{norm}}}{T}\sum_{t=1}^T g^{(t)} + \frac{\eta C^2}{2} + \frac{\lVert u\rVert_2^2}{2\eta T}.
\end{align*}
Bounding $g^{(t)}$ by $g_{\max}$ gives us
\begin{align*}
& \leq \frac{1}{T} \sum_{t=1}^T  \left\vert \ln\left(1 + e^{-y^{(t)}\tilde y^{(t)}}\right) - \ln\left(1 + e^{-y^{(t)} \hat y^{(t)}}\right)\right\vert  + g_{\max}\epsilon_{\text{norm}} + \frac{\eta C^2}{2} + \frac{\lVert u\rVert_2^2}{2\eta T}.
\end{align*}
Finally, by applying the inequality $\ln (1+x)\leq x \text{ for } x > -1$ and leveraging Lipschitz continuity, as well as Lemma~\ref{lem:quantum_inner_product_estimation}, we obtain:
\begin{align*}
& \leq \frac{1}{T}\sum_{t=1}^T e \epsilon_{\text{IP}} + g_{\max}\epsilon_{\text{norm}} + \frac{\eta C^2}{2} + \frac{\lVert u\rVert_2^2}{2\eta T}\\
& = e \epsilon_{\text{IP}} + g_{\max}\epsilon_{\text{norm}} + \frac{\eta C^2}{2} + \frac{\lVert u\rVert_2^2}{2\eta T}.
\end{align*}

Setting $\epsilon_{\text{IP}} = \frac{1}{4\eta T}$, $\epsilon_{\text{norm}} = \frac{1}{2\eta T}$ and $\eta = \frac{1}{C^2 \sqrt T}$, the desired regret bound is achieved with success probability $1-\delta$ by the union bound when quantum inner product and norm estimation each succeeds with probability $1-\delta/2$. 

Finally, we compute the total time complexity of the algorithm. The algorithm uses the following subroutines, each with the correspoinding running time:
\be\nonumber 
 T\left(\text{quantum inner product estimation + quantum norm estimation + quantum state preparation}\right)\\\nonumber
 \subseteq  O\left(T\left(\frac{T\sqrt d}{\epsilon_{IP}} \log\frac{T}{\delta}+ \frac{T\sqrt{d}}{\epsilon_{\text{norm}}} \log\frac{T}{\delta}+ T\sqrt{d}\log \frac{T}{\delta}\right)\right) \subseteq  O\left(T^{5/2}\sqrt{d}\log\frac{T}{\delta}\right).
\ee

\subsection{Proof of Theorem~\ref{theorem:regret_SVM}}\label{app:proof_regret_SVM}
We start by considering the following expression for the regret bound:
\begin{align*}
& \frac{1}{T}\sum_{t=1}^T \left(1-y^{(t)}\tilde y^{(t)}\right)^+ + \frac{1}{T}\sum_{t=1}^T g^{(t)}\tilde q^{(t+1)}\\
& -\frac{1}{T}\sum_{t=1}^T \left(1 -y^{(t)}u^T x^{(t)}\right)^+ - \frac{1}{T}\sum_{t=1}^T g^{(t)}\left\Vert u\cdot I\left(\left\vert w^{(t+1)}\right\vert\right)\right\Vert_1.
\end{align*}
Next, we separate the terms for terms for $\hat y^{(t)}$ and $\tilde y^{(t)}$ and apply Fact~\ref{fact:classical_regret}:
\begin{align*}
& \leq \frac{1}{T}\sum_{t=1}^T \left(1-y^{(t)}\tilde y^{(t)}\right)^+ + \frac{1}{T}\sum_{t=1}^T g^{(t)}\tilde q^{(t+1)}\\
& \qquad-\frac{1}{T}\sum_{t=1}^T \left(1-y^{(t)}\hat y^{(t)}\right)^+ - \frac{1}{T}\sum_{t=1}^T g^{(t)} q^{(t+1)} +\frac{C^2\lVert u\rVert_2^2}{2 \sqrt T} + \frac{1}{2\sqrt T}\\
& \leq \frac{1}{T}\sum_{t=1}^T \left\vert \left(1-y^{(t)}\tilde y^{(t)}\right)^+ - \left(1-y^{(t)}\hat y^{(t)}\right)^+\right\vert \\
& \qquad+ \frac{1}{T}\sum_{t=1}^T g^{(t)}\left\vert \tilde q^{(t+1)} -q^{(t+1)} \right\vert +\frac{C^2\lVert u\rVert_2^2}{2 \sqrt T} + \frac{1}{2\sqrt T}.
\end{align*}
Using additive variant of Lemma~\ref{lem:quantum_norm_est_and_quantum_state_prep}(\ref{lem:quantum_norm_estimation_multiplicative}) and $g_{\max} = \max_{t \in [T]} g^{(t)}$, we obtain:
\begin{align*}
& \leq \frac{1}{T}\sum_{t=1}^T \left\vert \left(1-y^{(t)}\tilde y^{(t)}\right)^+ - \left(1-y^{(t)}\hat y^{(t)}\right)^+\right\vert + \epsilon_{\text{norm}}g_{\max} +\frac{C^2\lVert u\rVert_2^2}{2 \sqrt T} + \frac{1}{2\sqrt T}.
\end{align*}
We finally obtain the bound by using Lipschitz continuity and Lemma~\ref{lem:quantum_inner_product_estimation}:
\begin{align*}
& \leq \vert \hat y^{(t)} - \tilde y^{(t)}\vert + \epsilon_{\text{norm}}g_{\max} + \frac{C^2\lVert u\rVert_2^2}{2 \sqrt T} + \frac{1}{2\sqrt T}\\
& \leq \epsilon_{\text{IP}} + \epsilon_{\text{norm}}g_{\max} +\frac{C^2\lVert u\rVert_2^2}{2 \sqrt T} + \frac{1}{2\sqrt T}\\
& \leq \frac{2 + C^2\left(g_{\max} + \Vert u\Vert_2^2\right)}{2\sqrt T}
\end{align*}
when setting $\epsilon_{\text{IP}} = \frac{1}{2\sqrt T}$, $\epsilon_{\text{norm}} = \frac{1}{2\eta T}$ and $\eta = \frac{1}{C^2\sqrt T}$. This succeeds with probability $1-\delta$ by the union bound when quantum inner product and norm estimation each succeeds with probability $1-\delta/2$. 

The time complexity analysis is the same as that of Theorem~\ref{lem:regret}. 

\subsection{Proof of Theorem~\ref{thm: QSOL_LS}}\label{app:proof_QSOL_LS}
We start by considering the following expression for the regret bound:
\begin{align*}
& \frac{1-2C^2\eta}{T}\sum_{t=1}^T\left(\tilde y^{(t)} - y^{(t)}\right)^2 + \frac{1}{T}\sum_{t=1}^T g^{(t)}\tilde q^{(t+1)}\\
& - \frac{1}{T}\sum_{t=1}^T \left(u^T x^{(t)} - y^{(t)}\right)^2 - \frac{1}{T}\sum_{t=1}^T g^{(t)}\left\Vert u\cdot I\left(\left\vert w^{(t+1)}\right\vert\leq\theta\right)\right\Vert_1.
  \end{align*}
At this point, we apply Fact~\ref{fact:classical_regret} to obtain the following bound:
\begin{align*}
& \leq \frac{1-2C^2\eta}{T}\sum_{t=1}^T\left(\tilde y^{(t)} - y^{(t)}\right)^2 + \frac{1}{T}\sum_{t=1}^T g^{(t)}\tilde q^{(t+1)}\\
& \qquad - \frac{1-2C^2\eta}{T}\sum_{t=1}^T\left(\hat y^{(t)} - y^{(t)}\right)^2 - \frac{1}{T}\sum_{t=1}^T g^{(t)}\left\Vert w^{(t+1)}\cdot I\left(\left\vert w^{(t+1)}\right\vert\leq\theta\right)\right\Vert_1 + \frac{\lVert u\rVert_2^2}{2\eta T}.
\end{align*}
Next, we separate the terms to focus on the error:
\begin{align*}
&\leq \frac{1-2C^2\eta}{T}\sum_{t=1}^T \left\vert \left(\tilde y^{(t)} - y^{(t)}\right)^2 - \left(\hat y^{(t)} - y^{(t)}\right)^2\right\vert  + \frac{1}{T}\sum_{t=1}^T g^{(t)}\left\vert\tilde q^{(t+1)} - q^{(t+1)}\right\vert + \frac{\lVert u\rVert_2^2}{2\eta T}.
\end{align*}
Use Lipschitz continuity to further bound it:
\begin{align*}
& \leq \frac{(1-2C^2\eta)\epsilon_{\text{IP}}}{T}\sum_{t=1}^T 2\left\vert \hat y^{(t)}  - y^{(t)}\right\vert\left\Vert x^{(t)}\right\Vert_2 + \frac{1}{T}\sum_{t=1}^T g^{(t)}\left\vert\tilde q^{(t+1)} - q^{(t+1)}\right\vert + \frac{\lVert u\rVert_2^2}{2\eta T}.
\end{align*}
By additive variant of Lemma~\ref{lem:quantum_norm_est_and_quantum_state_prep}(\ref{lem:quantum_norm_estimation_multiplicative}), now we have:
\begin{align*} 
& \leq \frac{(1-2C^2\eta)\epsilon_{\text{IP}}}{T}\sum_{t=1}^T 2\left\vert \hat y^{(t) } - y^{(t)}\right\vert\left\Vert x^{(t)}\right\Vert_2 + \frac{1}{T}\sum_{t=1}^T g^{(t)}\epsilon_{\text{norm}} + \frac{\lVert u\rVert_1}{2\eta T}.
\end{align*}
Applying Assumption~\ref{ass: regret_ass}(\ref{ass:sup_x}), we can simplify it:
\begin{align*}
& \leq \frac{2C\epsilon_{\text{IP}}(1-2C^2\eta)}{T}\sum_{t=1}^T \left\vert\hat y^{(t) } - y^{(t)}\right\vert + \frac{1}{T}\sum_{t=1}^T g^{(t)}\epsilon_{\text{norm}} + \frac{\lVert u\rVert_2^2}{2\eta T}.
\end{align*}
Finally, we apply $g_{\max} = \max_{t \in [T]} g^{(t)}$ and Assumption~\ref{ass:bound_D} to obtain:
\begin{align*}
& \leq \frac{2C\epsilon_{\text{IP}}(1-2C^2\eta)}{T}\sum_{t=1}^T \left\vert \hat y^{(t) } - y^{(t)}\right\vert + \frac{\epsilon_{\text{norm}}}{T}\sum_{t=1}^T g_{\max } + \frac{\lVert u\rVert_2^2}{2\eta T}\\
& \leq 2C\epsilon_{\text{IP}}(1-2C^2\eta) D + \epsilon_{\text{norm}} g_{\max} + \frac{\lVert u\rVert_2^2}{2\eta T}\\
& \leq 2CD\epsilon_{\text{IP}} + \epsilon_{\text{norm}} g_{\max} + \frac{\lVert u\rVert_2^2}{2\eta T}
\end{align*}
with success probability $1-\delta$ by the union bound when quantum inner product and norm estimation each succeeds with probability $1-\delta/2$. By setting $\epsilon_{\text{IP}} = \frac{1}{4\eta T}$, $\epsilon_{\text{norm}} = \frac{1}{2\eta T}$ and $\eta = \frac{1}{C^2\sqrt{T}}$, we obtain the desired bound. 

The time complexity analysis is the same as that of Theorem~\ref{lem:regret}. 

\end{document}